\documentclass{article}

\usepackage[preprint]{neurips_2026}

\usepackage[utf8]{inputenc}
\usepackage[T1]{fontenc}
\usepackage{hyperref}
\hypersetup{hidelinks}
\usepackage{url}
\usepackage{booktabs}
\usepackage{amsfonts}
\usepackage{amsmath}
\usepackage{nicefrac}
\usepackage{microtype}
\usepackage{xcolor}
\usepackage{graphicx}

\title{What pass@k Cannot Measure: Evaluating Diversity and Capability Retention after
Post-Training}

\author{
  Subham Rath \\
  \small Accenture Strategy \& Consulting \\[6pt]
  Raj Dandekar$^{1}$ \quad Rajat Dandekar$^{1}$ \quad Sreedath Panat$^{2}$ \\
  \small $^{1}$Vizuara AI Labs (vizuara.com) \quad $^{2}$Vizuara AI Labs (vizuara.ai)
}

\begin{document}

\maketitle

\begingroup
\footnotesize\centering
This paper was accepted to the NeurIPS 2026 Workshop on Transitioning from Pre-training to
Post-training. A shorter, evaluation-focused companion version was accepted at the JUDGe 2026
workshop (Can We Trust the Judge?), scheduled for presentation on December 11, 2026. Both are
non-archival.\par
\endgroup
\vspace{0.5em}

\begin{abstract}
pass@$k$, the fraction of problems a model solves within $k$ sampled attempts, is the field's
default protocol for deciding whether reinforcement-learning (RL) post-training on verifiable
rewards improved a model. At the population level, pass@$k$ depends only on a problem's
probability of a correct sample, with no term for how that probability is distributed across
distinct outputs. We show this structural gap is not academic. Training Qwen2.5-1.5B-Instruct
on grade-school math with Group Relative Policy Optimization (GRPO) and with rejection-sampling
fine-tuning (RFT, training on the model's own shortest verifier-passed rollout) moves three
complementary diversity measures (token-level entropy, answer-level entropy, and unique answers
per prompt) in opposite directions, with zero overlap across three seeds per arm. The gap
survives restricting to verifier-correct completions only (lexical diversity among correct
solutions is 15\% lower for GRPO, holding after controlling for completion length) and a
count-controlled check that isolates diversity among incorrect answers alone, ruling out that
the raw gap is just an artifact of GRPO's higher accuracy shrinking its incorrect-answer pool.
Yet pass@8 and pass@32 show no consistent winner on GSM8K, and a deliberately hard MATH-500
subset shows a related pattern: the arms separate only at low $k$ and converge as $k$ grows.
Compared against the starting checkpoint, no trained arm significantly improves hard-problem
coverage: RFT is significantly worse, while GRPO is statistically indistinguishable from the
starting checkpoint -- so GRPO's pass@1 edge over RFT there reflects a smaller loss relative to
Base, not a capability gain over it, a missing-control problem distinct from any failure of
pass@$k$ itself. On GSM8K, only pass@1, with no role in detecting diversity by construction,
separates the arms cleanly, and it rewards the arm whose correct solutions are the least
diverse. We argue this is a concrete instance of a standard evaluation protocol missing a
property it is routinely used to certify.
\end{abstract}

\section{Introduction}
\label{sec:intro}

Post-training is supposed to convert a broad-coverage base model into one that reliably solves
the tasks it was tuned for, without secretly narrowing what the model can still do.
Reinforcement learning with verifiable rewards (RLVR) is now standard practice for this
conversion on reasoning tasks: a deterministic correctness check replaces a learned reward
model, and training is judged almost entirely by pass@$k$ \citep{chen2021codex}: the probability
that at least one of $k$ independently sampled attempts at a problem is correct, computed with
the unbiased estimator $1 - \binom{n-c}{k}/\binom{n}{k}$ over $n \geq k$ samples with $c$
correct. pass@1 is treated as capability. pass@$k$ at larger $k$ is treated, implicitly, as a
check on whether that capability cost the model's solution diversity: a model that merely
rehearses one answer more confidently should still show it at large $k$, while a model that
discarded alternative correct paths will not.

This intuition is itself a construct-validity claim, and it is worth stating precisely why it
can fail. At the population level, for a problem with true per-sample correctness probability
$p$, pass@$k(p) = 1-(1-p)^k$, a function of $p$ alone. Two models with the same $p$ have the same
pass@$k$ at every $k$, regardless of how the remaining probability mass is arranged across
distinct correct solutions, distinct incorrect ones, or reasoning paths that reach the same
answer. pass@$k$ is a statement about how much correct mass exists, not about how it is
distributed. A post-training method that redistributes mass among already-correct outputs, or
among incorrect ones, without changing $p$ much, is therefore invisible to it by construction
rather than by an implementation gap. Whether this is the whole explanation for what follows is
an empirical question we return to in \S\ref{sec:discussion}. The point here is that the
construct-validity gap has a structural reason to exist, and is not only something we happened
to observe.

\citet{yue2025rl} gave the first large-scale evidence for part of this picture: RLVR-trained
models beat their base model at small $k$ but are overtaken once $k$ reaches the tens to
hundreds, across six RL algorithms and model families at 7B+ parameters, a pattern consistent
with post-training sharpening some capabilities while suppressing others rather than uniformly
broadening them. Raising an RL-trained model's sampling temperature to match the base model's
entropy narrows the pass@$k$ gap but does not close it (\citealp{yue2025rl}, \S4.5), so entropy
is necessary but not sufficient to explain it, suggesting entropy and pass@$k$ coverage are not
the same axis.

The literature disagrees on which post-training method causes the narrowing, and even on its
direction. \citet{nicolicioiu2026sdsd} find that on-policy self-distillation (training on one's
own sampled demonstrations) raises pass@1 while flattening pass@16, through a mechanism distinct
from ordinary RL mode-seeking; in one of their settings, self-distillation has \emph{higher}
token entropy than RL despite \emph{lower} pass@16. \citet{kim2026sdpo} find the opposite sign
for RL specifically: GRPO \emph{increases} token entropy on two reasoning-distilled 7--8B
models, while their self-distillation method (SDPO) decreases it, the reverse of what we report
below for GRPO at a much smaller scale on easier problems. This is a genuine, page-cited
disagreement: whether GRPO raises or lowers output entropy is unsettled, and appears to depend
on scale, starting checkpoint, and task difficulty in ways neither paper isolates.

Recent work addresses part of this gap directly: \citet{lee2026quatro} introduce UCC@k, which
clusters correct completions by semantic similarity and sums per-cluster coverage, so a model
that spreads correct mass across many distinct solutions scores higher than one that does not,
even at matched pass@$k$. Our question is complementary: when two post-training methods produce
reliably different changes across simpler, already-standard measures of the output distribution
(entropy, unique-answer counts) on the same starting checkpoint, does the field's existing,
unmodified pass@$k$ protocol register that difference, and what does comparison against the
starting checkpoint reveal about what post-training preserved? At a scale and difficulty level
neither \citet{yue2025rl}, \citet{nicolicioiu2026sdsd}, nor \citet{kim2026sdpo} tested
(Qwen2.5-1.5B-Instruct, grade-school arithmetic), our answer is no. This reframes the
contribution from "RL narrows diversity" (already claimed and disputed) to a claim about the
protocol itself: pass@$k$ at the $k$ values the field reports is not a valid standalone proxy for
that construct, independent of which direction any given method moves it. This bears directly on
evaluating post-training: an improved pass@$k$ alone cannot certify that post-training preserved,
rather than narrowed, what the starting checkpoint could do.

\section{Setup}
\label{sec:setup}

\paragraph{Model and data.} Qwen2.5-1.5B-Instruct is the pre-post-training baseline checkpoint
for every arm below, already instruction-tuned by its original authors; we do not train it
further until each arm's own run starts. Each arm trains on a frozen pool of 1{,}500 GSM8K
\citep{cobbe2021gsm8k} problems, filtered once against this checkpoint to keep only problems
where 1--7 of 8 sampled rollouts land correct (a zero-variance group produces zero GRPO
advantage and a degenerate RFT target, so these are excluded once, before training). Evaluation
is in-domain on a held-out 100-problem GSM8K set (32 samples/problem) and out-of-domain on
MATH-500 \citep{lightman2023math500} (16 samples/problem, never trained on).

\paragraph{Arms.} \textbf{GRPO} \citep{shao2024deepseekmath}: group-relative advantage over 8
rollouts/prompt, a standard clipped policy-gradient update, no KL penalty. \textbf{RFT}
(rejection-sampling fine-tuning): 8 rollouts/prompt, keep only the shortest verifier-passed one,
fine-tuned with plain cross-entropy on that single target, no distillation loss and no teacher
distribution, the simplest possible instance of training on one's own correct output. A second
RFT variant adds a Kullback-Leibler penalty toward the frozen starting policy (RFT + KL brake).
Both RFT arms train on model-generated, verifier-passed targets only, never the dataset's
reference solution. These three arms are the main comparison, each run at three seeds (0, 1, 2),
187 steps each. Three further arms run at a single seed for context, not a robustness claim: two
GRPO variants with a length penalty added to the reward ($\lambda = 0.2, 0.5$), and on-policy
self-distillation (OPD), a per-token reverse Kullback-Leibler divergence toward a
hint-conditioned copy of the same model. Every generation call, in training and evaluation,
uses temperature $1.0$, top-$p$ $1.0$, no top-$k$ truncation, and a $1{,}024$-token completion
budget; evaluation reuses the training-time top-$p$ and top-$k$, so no eval-time decoding
difference confounds a paper whose central claim is about entropy.

\paragraph{Metrics.} Token-level entropy, answer-level entropy, and unique answers per prompt are
all computed on the policy model's own tokenizer output, never a third-party tokenizer, alongside
pass@1/8/32. Each seed's own pass@$k$ carries a 95\% percentile bootstrap confidence interval
over its 100 evaluation problems \citep{chen2021codex}, computed at eval time. The 3-seed
summaries reported below combine the three seeds' intervals with a DerSimonian--Laird
random-effects meta-analysis, which folds in between-seed variance rather than treating three
different trained models as exact repeats of one measurement.

Three further metrics guard against a specific alternative explanation: that rising answer or
token entropy, or a rising raw unique-answer count, could simply mean more kinds of wrong
answers, or more answers because fewer are correct and available to repeat. (i) Distinct-3
\citep{li2016distinct}, the fraction of unique word-trigrams pooled across a problem's
verifier-correct completions only, computed on GSM8K rollouts where available (\S\ref{sec:limits}
for coverage) and again after truncating every completion to a shared word budget to rule out a
completion-length confound. (ii) $D_{\text{incorrect}}$, the fraction of unique incorrect answers
among a problem's incorrect completions only, averaged over problems. Correct completions never
enter its computation, but the size of the remaining incorrect-answer pool still scales with
accuracy, so we also report a count-controlled variant that subsamples a fixed number of
incorrect completions per problem (\S\ref{sec:results}). (iii) pass@$k$ for
$k \in \{1,2,4,8,16\}$ on MATH-500 restricted to its hardest levels (4--5, 100 problems, chosen
for headroom), generated freshly against each final checkpoint and against the same
pre-intervention checkpoint (3 sampling seeds of the same frozen weights, since nothing is
trained); the default pipeline only ever scored levels 1--2, so no rollouts existed here
beforehand. Arm-difference confidence intervals use a bootstrap clustered by problem: resample
the 100 problems, keep all 3 seeds or replicates per resample, and average the per-replicate
differences.

\section{Results}
\label{sec:results}

\begin{figure}[t]
\centering
\includegraphics[width=\linewidth]{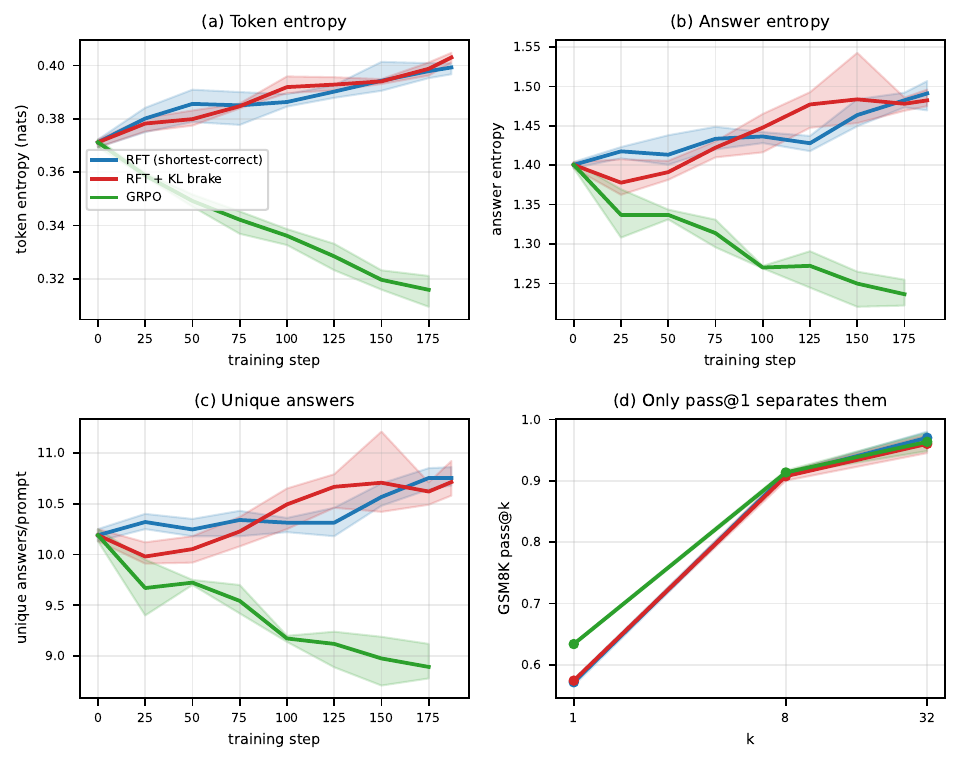}
\caption{GSM8K, all three arms, 3 seeds each; shaded bands span the seed min--max, not a CI.
\textbf{(a--c)} Token/answer entropy and unique answers over training: GRPO (green) and both
RFT arms (blue, red) move in opposite directions, with no band overlap from step 75 onward.
\textbf{(d)} Final-checkpoint pass@$k$: cleanly separated at $k{=}1$; no statistically detectable
separation at $k{=}8$ or $k{=}32$, same seeds and arms as (a--c).}
\label{fig:main}
\end{figure}

\paragraph{Diversity moves in opposite, seed-robust directions.} Figure~\ref{fig:main}(a--c) and
Table~\ref{tab:diversity} report the change in each diversity metric from the first to the last
evaluation checkpoint, as mean $\pm$ half-range across three seeds. Both RFT arms increase all
three measures in every seed, and GRPO decreases all three in every seed, with no seed of either
RFT arm close to zero or negative and no GRPO seed close to zero or positive. A mode-seeking
account would predict self-distillation-style training on one's own output as the mechanism
that narrows diversity; our result is the opposite, since GRPO narrows diversity here despite
being the arm least resembling self-distillation. One caveat applies to the
unique-answers-per-prompt column specifically: GSM8K has one canonical correct answer, so a
rising count could partly reflect an arm's falling accuracy, since more incorrect samples
mechanically allows more distinct values, rather than reflecting diversity on its own.
\S\ref{sec:results} below isolates diversity among incorrect answers only, to check whether the
direction survives once that mechanism is removed.

\begin{table}[t]
\caption{GSM8K diversity metrics, change from first to last checkpoint, mean $\pm$ half-range
across 3 seeds. All three move in the same direction within an arm, opposite across arms, with
zero cross-arm overlap in any seed.}
\label{tab:diversity}
\centering
\footnotesize
\renewcommand{\arraystretch}{0.82}
\begin{tabular}{lrrr}
\toprule
Arm & Token entropy $\Delta$ & Unique answers/problem $\Delta$ & Answer entropy $\Delta$ \\
\midrule
RFT (shortest-correct) & $+0.028 \pm 0.003$ & $+0.56 \pm 0.06$ & $+0.091 \pm 0.017$ \\
RFT + KL brake          & $+0.032 \pm 0.001$ & $+0.52 \pm 0.11$ & $+0.082 \pm 0.009$ \\
GRPO                    & $-0.055 \pm 0.004$ & $-1.30 \pm 0.20$ & $-0.164 \pm 0.018$ \\
\midrule
\multicolumn{4}{l}{\emph{Single seed each; context only, not a robustness claim:}} \\
GRPO + length penalty ($\lambda{=}0.2$) & $-0.055$ & $-1.26$ & $-0.163$ \\
GRPO + length penalty ($\lambda{=}0.5$) & $-0.050$ & $-1.20$ & $-0.150$ \\
On-policy self-distillation (OPD)       & $-0.003$ & $-0.34$ & $-0.035$ \\
\bottomrule
\end{tabular}
\end{table}

\paragraph{pass@1 separates the arms cleanly; pass@8/32 do not.}
Figure~\ref{fig:main}(d) and Table~\ref{tab:passk} report final-checkpoint pass@$k$: the 3-seed
DerSimonian--Laird pooled mean and 95\% CI for the reseeded arms (\S\ref{sec:setup}), one seed's
own bootstrap CI for the others. GRPO's pass@1 CI does not overlap either RFT arm's, while at
$k=8$ and $k=32$ all three overlap heavily. A direct paired bootstrap of the GRPO$-$RFT
difference, jointly resampling (problem, seed) units from the 2 seeds with intact raw GSM8K
rollouts (\S\ref{sec:limits}), sharpens this: the pass@1 gap is $+0.058$ [$0.045,0.071$] against
RFT-shortest and $+0.059$ [$0.046,0.073$] against RFT+brake, both excluding zero, while the
pass@8/32 gaps ($+0.005$ to $+0.020$) all include zero. Simulating this design at 80\% power and
$\alpha{=}0.05$ puts the minimum detectable pass@1 effect near $0.02$, well below the observed
gap; the same floor at pass@8/32 ($0.02$--$0.03$) means a real effect of that size could have
gone undetected.

\begin{figure}[t]
\centering
\includegraphics[width=\linewidth]{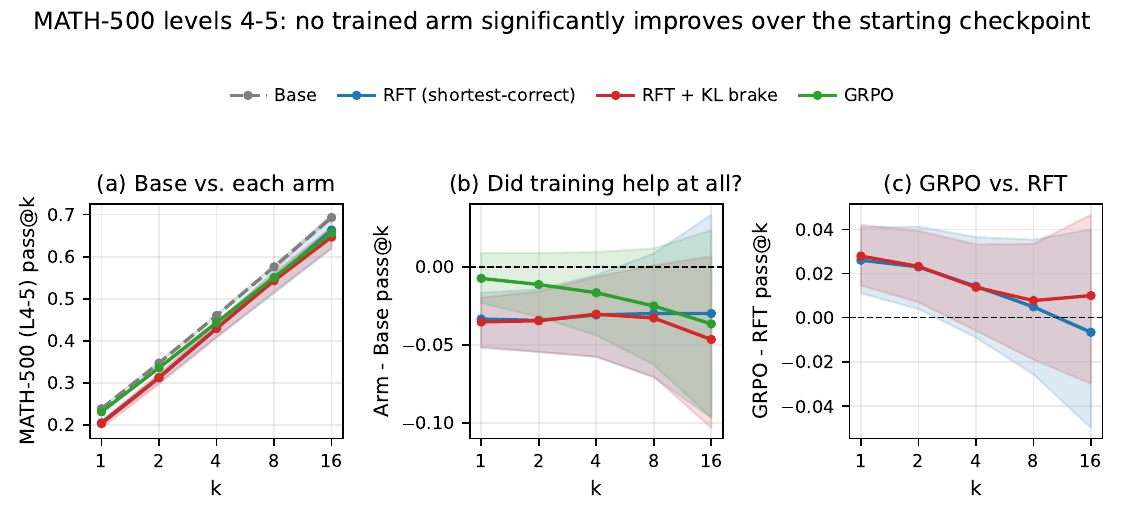}
\caption{MATH-500 levels 4--5 (100 problems, $n{=}16$/problem), chosen for headroom: pass@16
stays below $0.67$. Bands are min--max over 3 seeds (trained arms) or 3 sampling seeds of the
shared pre-intervention checkpoint (Base). \textbf{(a)} Base sits above every trained arm at
every $k$.
\textbf{(b)} Arm$-$Base, 95\% bootstrap CI clustered by problem: RFT is significantly below
baseline through $k{=}4$; GRPO never significantly differs from it. \textbf{(c)} GRPO$-$RFT,
using the same method; the interval excludes zero at $k{=}1,2$ and includes it from $k{=}4$ on.}
\label{fig:math500hard}
\end{figure}

A more informative comparison is against the shared pre-intervention baseline rather than
comparing trained arms only against one another, since it asks whether training earned its keep
rather than merely outperformed a different treatment. Both RFT arms land below baseline on
pass@1, in-domain and out-of-domain (GSM8K $\Delta$: $-0.019$ to $-0.008$; MATH-500 levels 1--2
$\Delta$: $-0.037$ to $-0.014$, all 6 RFT-family seeds), while GRPO's gain ($+0.048$ to $+0.052$
GSM8K, $+0.026$ to $+0.033$ MATH-500 levels 1--2) transfers to the never-trained-on benchmark in
every seed, on problems the baseline could already handle well. That gain does not survive on
levels 4--5 (\S\ref{sec:results}, Figure~\ref{fig:math500hard}b), where GRPO's pass@1 is
statistically indistinguishable from baseline.

\begin{table}[t]
\caption{GSM8K pass@$k$, final checkpoint. Top: 3-seed DerSimonian--Laird pooled mean, 95\% CI.
Bottom: single seed, own bootstrap CI. Only pass@1's CI separates cleanly across arms.}
\label{tab:passk}
\centering
\footnotesize
\renewcommand{\arraystretch}{0.82}
\begin{tabular}{lccc}
\toprule
Arm & pass@1 & pass@8 & pass@32 \\
\midrule
RFT (shortest-correct) & $0.571$ [$0.538$, $0.605$] & $0.910$ [$0.886$, $0.935$] & $0.972$ [$0.955$, $0.990$] \\
RFT + KL brake          & $0.574$ [$0.541$, $0.607$] & $0.908$ [$0.882$, $0.934$] & $0.967$ [$0.945$, $0.988$] \\
GRPO                    & $\mathbf{0.634}$ [$\mathbf{0.598}$, $\mathbf{0.669}$] & $0.914$ [$0.887$, $0.940$] & $0.969$ [$0.951$, $0.988$] \\
\midrule
GRPO + length penalty ($\lambda{=}0.2$) & $0.630$ [$0.565$, $0.687$] & $0.913$ [$0.865$, $0.954$] & $0.960$ [$0.920$, $0.990$] \\
GRPO + length penalty ($\lambda{=}0.5$) & $0.625$ [$0.560$, $0.682$] & $0.908$ [$0.858$, $0.950$] & $0.950$ [$0.900$, $0.990$] \\
OPD                                     & $0.599$ [$0.532$, $0.653$] & $0.911$ [$0.866$, $0.952$] & $0.970$ [$0.930$, $1.000$] \\
\bottomrule
\end{tabular}
\end{table}

\paragraph{The gap survives restricting to correct solutions only.} The sharpest objection is
that raw unique-answer and answer-entropy counts pool correct and incorrect completions, so
RFT's higher numbers could simply mean more kinds of wrong answers rather than richer correct
reasoning. Restricting to verifier-correct GSM8K completions and computing distinct-3, the
fraction of unique word-trigrams pooled across a problem's correct completions
\citep{li2016distinct}, GRPO's correct solutions are less diverse than RFT's: $0.436$ (GRPO)
versus $0.512$/$0.509$ (RFT arms), all $n{=}2$ seeds, a 15\% gap in the same direction as the
all-completions result, on the same pass@32 that cannot tell the arms apart. We could not
compute the entropy analogue restricted to correct rollouts, since only a batch-mean is logged
at eval time, never per-sample.

\paragraph{The correct-only gap is not a completion-length artifact.} RFT's correct completions
are shorter than GRPO's at the final checkpoint (mean tokens among verifier-correct completions:
$272.7$--$272.8$ versus $296.4$; median: 263--266 versus 285--288; both $n{=}2$ seeds),
consistent with training toward the shortest verifier-passed rollout. Distinct-3 is
length-sensitive: pooling more text gives trigrams more chances to repeat. Truncating every
correct completion to a shared word budget (157 words, the shorter arms' own median) before
recomputing distinct-3 gives $0.456$ (GRPO) versus $0.531$/$0.528$ (RFT arms), a $14\%$ gap,
within one point of the untruncated $15\%$.

\paragraph{Restricting to genuinely distinct wrong answers narrows the raw gap, but does not
close it.} A sharper version of the same objection: GSM8K has one canonical correct answer, so
GRPO's higher accuracy shrinks its incorrect pool, which can only reduce, never raise, how many
distinct wrong values it contains, so part of Table~\ref{tab:diversity}'s gap could be accuracy
passing through the metric. $D_{\text{incorrect}}$ (\S\ref{sec:setup}), computed entirely within
incorrect completions, removes this specific mechanism: $0.758$ (GRPO) versus $0.769$/$0.768$
(RFT arms), an order of magnitude smaller than the raw delta and a real correction rather than a
confirmation of it. This raw version still lets each arm's incorrect-pool size vary with its
accuracy, so we also count-control it (subsampling exactly 4 incorrect samples per problem where
at least 4 exist): the gap widens to $0.873$ versus $0.898$/$0.892$, a $2.2$--$2.9\%$ relative
gap; both versions are real but small, and most of the raw magnitude in
Table~\ref{tab:diversity} is accuracy passing through the metric. The correct-only distinct-3
gap ($14$--$15\%$) does not rely on the incorrect-answer pool at all and provides a
complementary correct-only check, which is why it remains the strongest single diversity
result.

\paragraph{A harder benchmark with real headroom shows the same loss of separation, ruling
out saturation as the explanation.} A standing objection to every result above is that GSM8K's
pass@$k$ is already near 0.9 by $k{=}8$ (Table~\ref{tab:passk}), so convergence past $k{=}8$
could reflect a benchmark ceiling rather than a construct-validity gap. MATH-500 restricted to
its hardest levels (4--5) is a stricter test of that objection: pass@1 is only $0.20$--$0.23$,
and pass@16 stays below $0.67$, leaving substantial headroom at every $k$ we test
(Figure~\ref{fig:math500hard}). GRPO exceeds both RFT variants at pass@1 and pass@2, with a
clustered bootstrap CI that excludes zero at both, but from $k{=}4$ on the CI includes zero for
both comparisons, a narrower claim than showing the methods equivalent. Because this happens far
short of the benchmark's own ceiling, the loss of separation cannot be explained solely by
saturation. The same pattern appears on the full MATH-500 set, which has less headroom; levels
4--5 give the stricter test.

\paragraph{No trained arm significantly improves over the starting checkpoint on hard
problems.} pass@1 correctly reports which treatment has more correct-answer mass; it does not by
itself say whether either helped relative to no intervention, which requires comparison with the
starting checkpoint. We evaluated the same starting checkpoint on the same 100 levels-4--5
problems, using 3 independent sampling seeds of the same frozen weights, since nothing is
trained (\S\ref{sec:setup}).
Figure~\ref{fig:math500hard}(a,b) shows the baseline sitting above every trained arm at every
$k$. A clustered bootstrap of arm minus baseline finds RFT significantly below baseline at
$k{=}1,2,4$ ($-3$ to $-3.5$pp for both variants, CI excludes zero), meaning training actively
cost coverage rather than merely failing to add it. GRPO never significantly differs from
baseline, though its point estimate drifts from $-0.7$pp at $k{=}1$ to $-3.7$pp at $k{=}16$.

Read together with panel (c), GRPO's win over RFT here reflects a smaller loss relative to the
starting checkpoint rather than a capability gain over it, matching the distinction
\citet{yue2025rl} draw between sharpening existing capability and adding new support
(\S\ref{sec:intro}), and the opposite of levels 1--2, where GRPO's gain over baseline is real
(above). This is also a different failure from the structural argument in \S\ref{sec:intro}:
pass@1 measured the right quantity, correctly finding more correct mass in GRPO than in RFT, and
the risk is entirely in reading a treatment gap as a capability claim with no control to check
it against.

\paragraph{The arm that is actually self-distillation-shaped moves least of all.} The literature
this paper engages with (\S\ref{sec:intro}) is about self-distillation; RFT's cross-entropy loss,
lacking a teacher distribution or divergence term, does not really belong to that family, but
OPD does. At one seed each, the length-penalized GRPO variants track plain GRPO closely on every
diversity metric (Table~\ref{tab:diversity}), while OPD is the outlier, nearly unchanged from
baseline on every diversity metric and on pass@1 ($0.593\to0.599$, GSM8K). This is consistent
with under-training at this step budget, but equally consistent with genuine diversity
preservation; one seed cannot distinguish the two. It does establish that OPD, this project's
actual self-distillation arm, does not reproduce \citet{kim2026sdpo}'s finding that
self-distillation sharply lowers entropy, at this scale. Separately, GSM8K truncation
($\leq\!0.0012$ movement) and format-failure ($0.0000$ throughout) across GRPO seeds rule out a
formatting artifact behind its pass@1 gain.

\section{Discussion}
\label{sec:discussion}

\paragraph{An improved pass@$k$ does not certify that post-training preserved what the base
model could do.} Practitioners read pass@1 as evidence that a post-training run worked, and
pass@$k$ at larger $k$, implicitly, as evidence that it did not cost the model anything else.
Here, that second reading fails even in a case with no evaluator to blame: the verifier and
scoring rule are deterministic, so any blind spot in pass@$k$ is a property of the metric
itself, not of noise in how it is scored. Even so, pass@$k$ still fails to certify a construct
that three complementary measures show moved by a large, seed-robust margin in opposite
directions for two post-training methods applied to the same starting checkpoint. pass@$k$ is a
statement about correct-answer mass; it says nothing about how post-training reorganized the
model's output distribution to produce that mass.

This paper makes two separate claims. The first is structural (\S\ref{sec:intro}): pass@$k$
cannot see how probability mass is distributed across outputs, regardless of what it is compared
against. The second is the baseline comparison in \S\ref{sec:results}, where pass@1 correctly
reports which treatment has more correct mass and the risk lies in reading that comparison
without a control; only the first claim concerns construct validity.

\paragraph{The hard-problem regression is consistent with an alignment tax.} On MATH-500
levels 4--5, RFT regresses significantly below the starting checkpoint through $k{=}4$, and GRPO
never significantly improves over it (\S\ref{sec:results}). Read as a post-training failure
mode, this is consistent with an alignment-tax interpretation, where capability outside the
training distribution degrades as a side effect of optimizing within it, although our
experiments do not establish diversity narrowing as its cause. The diversity narrowing we
measure elsewhere is a plausible, not causally established, mechanism for the same regression,
consistent with \citet{yue2025rl}'s boundary-narrowing result.

Across the RLVR literature, pass@$k$ at $k \in \{8,32\}$ is used as evidence for whether training
collapsed a model's reasoning options. Here, three diversity measures moved in opposite
directions for the two methods, detectable at pass@1--2 even on a hard, high-headroom benchmark,
while pass@8 and pass@32 on GSM8K, and pass@8 and pass@16 on the harder MATH-500 subset,
registered nothing beyond noise (\S\ref{sec:results}). A practitioner reading only those
larger-$k$ values would see no detectable separation past $k{=}1$--$2$; one reading entropy and
answer-diversity would conclude they are opposites. These observations are not contradictory;
they reflect different constructs. pass@$k$ has its own blind spots, and treating large-$k$
pass@$k$ as sufficient evidence about diversity can therefore lead to a construct-validity
error when evaluating whether a post-training run changed more than the benchmark score shows,
including in our own assumptions before this result. Which direction either method moves
diversity is disputed elsewhere at larger scales
\citep{nicolicioiu2026sdsd, kim2026sdpo}; our claim concerns the protocol used to evaluate
post-training, not which method is better.

\section{Limitations}
\label{sec:limits}

\textbf{Scale.} All results are at 1.5B parameters; whether the effect and the metric's
blindness to it hold at 7B+ is untested, and \citet{yue2025rl} find pass@$k$ already
near-saturated by $k=1$ at that scale. \textbf{Data availability.} A since-fixed logging bug
overwrote GSM8K's raw per-sample rollouts for the first seed of every arm (MATH-500's were
unaffected, \S\ref{sec:setup}), so correct-only distinct-3, length-matched distinct-3,
$D_{\text{incorrect}}$, and the paired arm-difference are $n{=}2$ seeds, weaker than the $n{=}3$
elsewhere. Table~\ref{tab:passk}'s 3-seed CIs use a DerSimonian--Laird meta-analysis of only
$k{=}3$ studies and should be read as approximate. \textbf{Confounded comparison.} RFT and GRPO
differ on more than one axis at once (loss function, whether incorrect
rollouts enter the gradient, a group-relative baseline), so this paper reports what the metric
registered without identifying which mechanistic difference is responsible. \textbf{Single
starting checkpoint.} Every arm post-trains the same Qwen2.5-1.5B-Instruct checkpoint; we vary
post-training method, not base-model family, pretraining data, or starting-checkpoint
properties, so this paper does not speak to what a base model must possess for post-training to
succeed, only to how two post-training methods reorganize one particular starting point
differently.

\section*{Acknowledgments}
Experiments were run on Modal serverless GPU infrastructure. Generative AI tools were used for
language editing, code assistance, and figure generation during manuscript preparation. The
authors determined the research questions, experimental design, statistical methodology,
interpretation, and conclusions, and independently verified the reported results and
citations.

\bibliographystyle{plainnat}
\bibliography{judge_refs}

\newpage
\section*{Appendix: Implementation and Reproducibility Details}
\label{sec:appendix}

This appendix states only what is directly verifiable from this project's saved run
configurations (one \texttt{config.resolved.yaml} per run, committed alongside the training and
analysis code) and the source itself. It is not a claim that every number in the paper can be
reproduced exactly from released artifacts: the configs, scripts, and rollouts referenced below
are not (yet) packaged for public release alongside this preprint.

\paragraph{Training.} All arms fully fine-tune Qwen2.5-1.5B-Instruct in bfloat16, with no
LoRA or other adapter layers; the optimizer is AdamW. \textbf{GRPO:} learning rate
$1\times10^{-6}$, clip $\epsilon = 0.2$, KL coefficient $\beta = 0$ (no reference-model KL
penalty), 8 rollouts/prompt, 8 prompts/batch, gradient accumulation over 8 steps, per-device
batch size 8, 1 update per batch, 1 epoch over 1{,}500 training prompts. \textbf{RFT:} learning
rate $1\times10^{-6}$, plain cross-entropy on the single shortest verifier-passed rollout per
prompt, the same rollout/batch/prompt counts as GRPO.

\paragraph{Sampling and evaluation.} Generation temperature 1.0, top-$p$ 1.0, top-$k$
unrestricted; maximum prompt and completion length 1{,}024 tokens each. In-domain (GSM8K)
evaluation draws 32 samples/problem; out-of-domain (MATH-500) draws 16 samples/problem;
checkpoints are evaluated every 25 training steps.

\paragraph{Verifier and answer normalization.} Correctness is checked by a dedicated
normalizer (\texttt{src/opsd/verify/math\_verify.py}) that strips LaTeX wrapper commands
(\texttt{\textbackslash left}, \texttt{\textbackslash text}, \texttt{\textbackslash frac},
\texttt{\textbackslash sqrt}, etc.), currency/percent symbols, thousands separators, and
boilerplate phrasing such as ``the answer is''; lower-cases the result; and compares numerically
as an exact rational when both sides parse as numbers, falling back to normalized-string
equality otherwise. The same function is used for the training reward and for evaluation
scoring.

\paragraph{Entropy definitions.} Both entropies are Shannon entropy in nats
(\texttt{src/opsd/eval/diversity.py}). Token entropy at a generated position is
$H_t = -\sum_v p_t(v)\log p_t(v)$ over the full next-token vocabulary (not a sampled estimate),
averaged over generated positions and over the evaluation batch. Answer entropy for a prompt is
the Shannon entropy of the categorical distribution of \emph{normalized} final answers across
that prompt's samples; an unparseable completion counts as its own category rather than being
dropped, so a model degenerating into format failures is not misread as more concentrated.

\paragraph{Distinct-3 and $D_{\text{incorrect}}$.} Distinct-3 \citep{li2016distinct} is the
fraction of unique word-trigrams pooled across a problem's completions, computed both over all
completions and, separately, over verifier-correct completions only.
$D_{\text{incorrect}}(\text{problem}) = \lvert\text{unique incorrect answers}\rvert /
\lvert\text{incorrect completions}\rvert$, averaged over problems; correct completions never
enter this computation.

\paragraph{Bootstrap and statistical procedure, by analysis.} These are not one shared
setting. \textbf{(i)} The GSM8K pass@16 recomputation and the per-seed pass@$k$ confidence
intervals shown throughout use the training pipeline's own unbiased estimator and percentile
bootstrap (\texttt{src/opsd/eval/passk.py}, functions \texttt{pass\_at\_k\_batch} and
\texttt{bootstrap\_ci}), 1{,}000 resamples, seed 0. \textbf{(ii)} The MATH-500 hard-level
(levels 4--5) arm-versus-baseline clustered-by-problem bootstrap uses 3{,}000 resamples, seed
0. \textbf{(iii)} $D_{\text{incorrect}}$ and the count-controlled incorrect-only checks use
2{,}000 resamples, seed 0; the separate minimum-detectable-effect/power calculation reported in
the abstract uses 1{,}500 resamples, seed 1 (deliberately a different seed, independent of the
point-estimate bootstrap). \textbf{(iv)} The 3-seed pooled pass@$k$ summaries use a closed-form
DerSimonian--Laird random-effects meta-analysis over each seed's own point estimate and standard
error, which is not a resampling bootstrap.

\paragraph{Seeds and run structure.} Each trained arm has 3 independent training seeds. A
since-fixed logging bug overwrote GSM8K's raw per-sample rollouts for seed 0 of every arm
(MATH-500 was unaffected); the correct-only distinct-3, length-matched distinct-3,
$D_{\text{incorrect}}$, and paired arm-difference results are therefore $n=2$ seeds rather than
$n=3$, as stated in \S\ref{sec:limits}.

\paragraph{Known gap.} Per-rollout token entropy restricted to correct-only completions cannot
be reconstructed from logged data: the training/eval pipeline stores only a batch-mean token
entropy at evaluation time, never a per-sample value to filter by correctness. This is stated
where it matters, in \S\ref{sec:results}, rather than approximated here.

\end{document}